\documentclass[letterpaper, 10 pt, conference]{ieeeconf}  

\IEEEoverridecommandlockouts                              

\usepackage{graphics} 
\usepackage{epsfig} 
\usepackage{amsmath} 
\usepackage{amssymb}  
\usepackage{lipsum}
\usepackage{bm}
\usepackage{upgreek}
\usepackage{xcolor}
\usepackage{subfigure}
\usepackage{nicefrac}

\makeatletter
\@ifundefined{laTeXML@version}{}{%
  \let\DeclareSourcemap\@gobble
  \let\maps\@gobble
  \let\map\@gobble
  \let\pertype\@gobble
  \let\step\@gobble
}
\makeatother
\usepackage[style=ieee]{biblatex}
\usepackage[style=ieee]{biblatex}

\DeclareSourcemap{
  \maps{
    \map{
      \pertype{article}
      \step[fieldset=language, null]
      \step[fieldset=url, null]
      \step[fieldset=doi, null]
      \step[fieldset=issn, null]
      \step[fieldset=isbn, null]
      \step[fieldset=note, null]
      \step[fieldset=editor, null]
      \step[fieldset=urldate, null]
      \step[fieldset=file, null]
    }
  }
}
\DeclareSourcemap{
  \maps{
    \map{
      \pertype{inproceedings}
      \step[fieldset=language, null]
      \step[fieldset=url, null]
      \step[fieldset=doi, null]
      \step[fieldset=issn, null]
      \step[fieldset=isbn, null]
      \step[fieldset=note, null]
      \step[fieldset=editor, null]
      \step[fieldset=urldate, null]
      \step[fieldset=file, null]
    }
  }
}
\DeclareSourcemap{
  \maps{
    \map{
      \pertype{incollection}
      \step[fieldset=language, null]
      \step[fieldset=url, null]
      \step[fieldset=doi, null]
      \step[fieldset=issn, null]
      \step[fieldset=isbn, null]
      \step[fieldset=note, null]
      \step[fieldset=editor, null]
      \step[fieldset=urldate, null]
      \step[fieldset=file, null]
    }
  }
}

\newcommand{\mycdots}{\makebox[0.7em][c]{\!.\!\!\!.\!\!\!.}}

\title{\LARGE \bf
Quadratic Programming-Based Posture Manipulation and Thrust-vectoring for Agile Dynamic Walking on Narrow Pathways
}

\author{Chenghao Wang$^{1}$, Eric Sihite$^{2}$, Kaushik Venkatesh Krishnamurthy$^{1}$, Shreyansh Pitroda$^{1}$, \\ Adarsh Salagame$^{1}$, Alireza Ramezani$^{1*}$, and Morteza Gharib$^{2}$
\thanks{$^{1}$ The author is with Department of Electrical and Computer Engineering, Northeastern University, Boston, MA, USA  { wang.chengh, venkateshkrishnamu.k, pitroda.s, a.salagame, a.ramezani@northeastern.edu}}%
\thanks{$^{2}$ The author is with the Department of Aerospace Engineering, California Institute of Technology, Pasadena, CA, USA { esihite, mgharib@caltech.edu}}%
\thanks{$^{*}$ Corresponding author. Email: {a.ramezani@northeastern.edu}}%
}

\begin{document}

\maketitle
\thispagestyle{empty}
\pagestyle{empty}

\begin{abstract}

There has been significant advancement in legged robot's agility where they can show impressive acrobatic maneuvers, such as parkour. These maneuvers rely heavily on posture manipulation. To expand the stability and locomotion plasticity, we use the multi-modal ability in our legged-aerial platform, the Husky Beta, to perform thruster-assisted walking. This robot has thrusters on each of its sagittal knee joints which can be used to stabilize its frontal dynamic as it walks. In this work, we perform a simulation study of quadruped narrow-path walking with Husky $\beta$, where the robot will utilize its thrusters to stably walk on a narrow path. The controller is designed based on a centroidal dynamics model with thruster and foot ground contact forces as inputs. These inputs are regulated using a QP solver to be used in a model predictive control framework. In addition to narrow-path walking, we also perform a lateral push-recovery simulation to study how the thrusters can be used to stabilize the frontal dynamics.

\end{abstract}


\section{Introduction}

The quest for agile robot locomotion has driven significant advancements in legged locomotion, with numerous bipedal and quadrupedal robots showcasing impressive behaviors such as gymnastic routines and parkour \cite{noauthor_robots_nodate}.

These locomotion feats rely heavily on posture manipulation, which remains limited in terms of ground contact force manipulations. Birds \cite{abourachid_hoatzin_2019,gatesy_bipedal_1991}, which combine bipedalism and flight and are known for their impressive locomotion plasticity \cite{sihite_multi-modal_2023}, utilize both posture manipulation and thrust-vectoring. Consequently, they are able to achieve agile locomotion feats far beyond those of standard legged systems. Examples include Chukar birds, which adeptly ascend steep inclines \cite{dial_wing-assisted_2003}, executing agile maneuvers such as rapid walking, leaping, and jumping using both their legs and wings.

\begin{figure}[t]
    \centering
    \includegraphics[width =1 \linewidth]{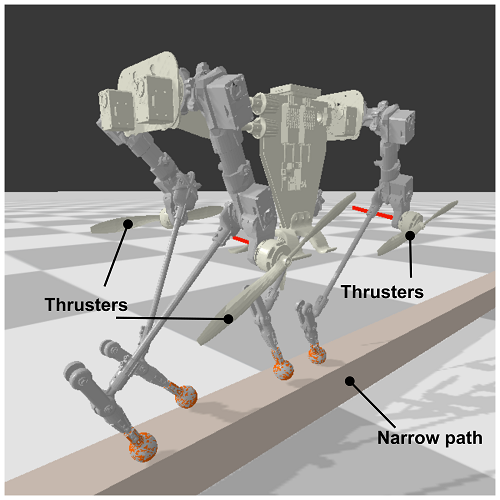}
    \caption{
     Northeastern University's Husky $\beta$ robot traversing a narrow beam uses sagittal thrust force and ground reaction force. The inward red line visualizes the direction and magnitude of the thrust force. 
    }
    \label{fig:cover-image}
    \vspace{-0.5cm}
\end{figure}

This study is a sequel to our previous publication on control design for thruster-assisted walking of the Northeastern Husky \cite{ramezani_generative_2021,liang_rough-terrain_2021,sihite_unilateral_2021,sihite_optimization-free_2021,salagame_quadrupedal_2023,dangol_control_2021,dangol_hzd-based_2021,dangol_performance_2020,sihite_demonstrating_2023,pitroda_capture_2024,pitroda_enhanced_2024,pitroda_quadratic_2024}, illustrated in Fig.~\ref{fig:cover-image}. Specifically, in this work, we explore dynamic walking on a rigid pipe. This problem is interesting because foot placement is very limited and cannot support posture manipulation for contact force manipulations. We fixate external thruster forces at four knee locations, which aligns with Husky's design. These forces translate with respect to the body COM as the legs move. The key question is how to recruit these four external forces and joint accelerations to dynamically walk over a pipe. 
 
\begin{figure*}[t]
    \centering
    \includegraphics[width=\linewidth]{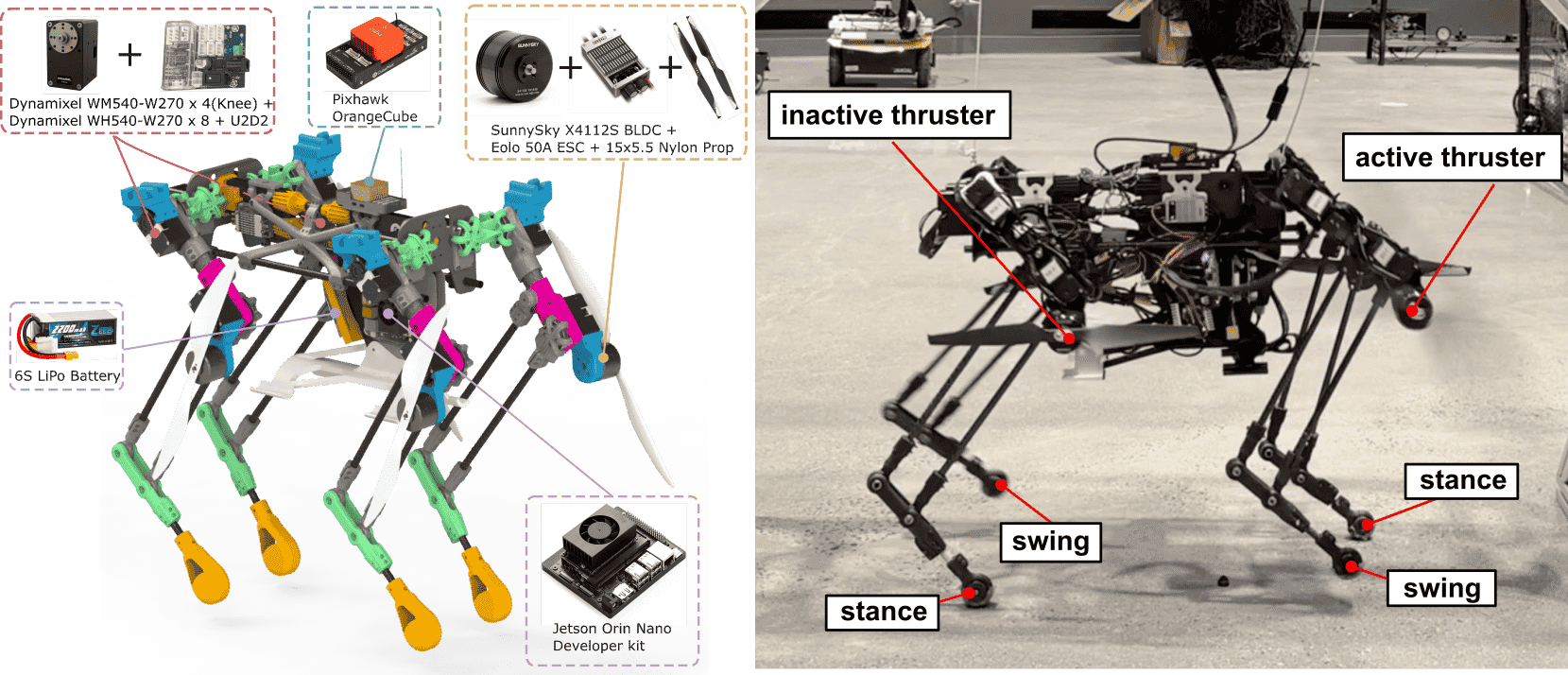}
    \caption{Hardware overview of Husky $\beta$ and its primary computing, power, and actuator components.}
    \label{fig:hardware}
\end{figure*}
Dynamic walking on a rigid pipe demands a control strategy that integrates posture adjustment with thrust vectoring, achieved via real-time optimization. In this context, several Python-based QP solvers prove essential. OSQP \cite{stellato_osqp_2020} offers robust infeasibility detection and real-time performance, while qpSWIFT \cite{pandala_qpswift_2019} excels at handling sparse problems. HiGHS \cite{huangfu_parallelizing_2018} suits large-scale models, and ECOS \cite{domahidi_ecos_2013} meets embedded systems’ needs with SOCP. Additionally, SCS \cite{odonoghue_operator_2020} efficiently addresses large-scale quadratically constrained problems, and both PROXQP \cite{bambade_proxqp_2023} and qpOASES \cite{ferreau_qpoases_2014} are effective in real-time applications—with PROXQP optimized for robotics. Recent benchmarks \cite{qpbenchmark2024} show that quadprog \cite{quadprog2021} offers the fastest runtime for small problems. These optimization tools serve as solvers for our MPC formulation, seamlessly linking our dynamic walking strategy with precise thruster-assisted control.


The primary goal of this research is to devise a control design methodology that integrates Husky's posture and its thrust vectoring capability, thereby enabling dynamic traversal of narrow paths. Our overarching objective is to achieve dynamic walking on a flexible rope, with the findings presented here marking a significant stride toward this ambition.

The structure of this work is as follows: 
{
a brief hardware overview of the Husky $\beta$ robot, followed by the centroidal dynamic modeling with thrusters, MPC formulation, simulation results and discussion, then finalized by the concluding remarks and future work.
}

\section{Overview of Northeastern Husky Platform}

The model used in this study is from the Husky robot family, which is being iteratively designed at Northeastern. The version utilized here is called version $\beta$ (shown in Fig.~\ref{fig:hardware}). In this design, a propeller motor is attached to the outside of each knee joint, differing from \cite{ramezani_generative_2021}, allowing the robot to morph into a quadrotor configuration by extending the hip frontal joints \cite{wang_legged_2023}. There are three actuated degrees of freedom per leg: hip frontal abduction-adduction, hip sagittal swing, and knee flexion-extension, similar to \cite{ramezani_generative_2021}.

To simplify the design of the robot, off-the-shelf servomotors are used to actuate each joint instead of the lighter, contrary to more specialized custom hardware pursued in \cite{ramezani_generative_2021}. The actuators are Dynamixel XH540-W270-T, which have a maximum stall torque of 9.9 Nm. To achieve flight capabilities, a combination of the SunnySky X4112S BLDC, EOLO 50A LIGHT ESC, and a 15x5.5 double-blade propeller was employed. This setup, with four such combinations, can generate a maximum thrust of 10.6 kg, providing a thruster-to-weight ratio of approximately 1.6, which is enough for flight tests. 
We developed a simulation of Husky-$\beta$ in PyBullet, as depicted in Fig.~\ref{fig:cover-image}.

\section{Centroidal Dynamic Model with Thrusters}
\label{sec:modeling}

Contrary to classical legged systems that use only ground reaction forces, the design paradigm of Husky enables lateral thrust forces generated by propellers as additional external forces. Thus, both ground reaction forces and thrust forces are incorporated into this centroidal dynamic model.

\subsection{Centroidal Dynamics Formulation}

\begin{figure}
    \centering
    \includegraphics[width=1\linewidth]{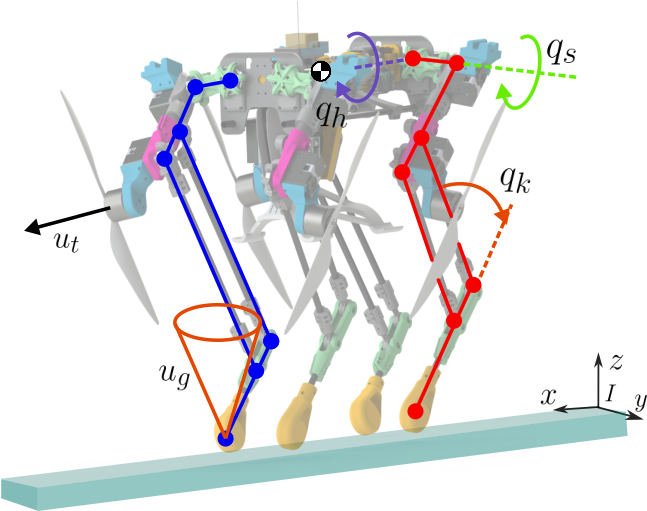}
    \caption{Free-body diagram of Husky-$\beta$ narrow path walking showing the inertial center of mass model and ground contact forces acting on the stance feet.}
    \label{fig:fbd}
\end{figure}

The ground reaction forces $\bm u_{g,i} \in \mathbb{R}^{3}$ and thrust forces $\bm u_{t,i} \in \mathbb{R}$ are applied on the center of mass through length vectors $\bm d_i \in \mathbb{R}^{3}$ and $\bm r_i \in \mathbb{R}^{3}$, respectively, as shown in Fig.~\ref{fig:fbd}. Since $\bm u_{t,i}$ can only be controlled in a single direction, an extra term $\hat{\bm e}_i \in \mathbb{R}^{3}$ is used to represent the orientation of thrusters. Let $\bm R \in SO(3)$ be a rotation matrix that rotates the body frame into the inertial frame. The dynamical state acceleration of the signal rigid body can be defined as follows:
\begin{equation}
\begin{aligned}
    \ddot{\bm p} &= \bm M^{-1} \left( 
         {\textstyle \sum_{i=1}^{4}} \left( \hat{\bm e}_i \, u_{t,i} + \bm u_{g,i} \right) \right)  +[0, 0, -g]^\top \\
\end{aligned}
\label{eq:dynamic_p}
\end{equation}
\begin{equation}
\begin{aligned}
    \frac{\mathrm{d}}{\mathrm{d} t}(\mathbf{I} \boldsymbol{\omega}) = 
    {\textstyle \sum_{i=1}^{4}} \left( \bm r_i \times \hat{\bm e}_i \, u_{t,i} + \bm d_i \times \bm u_{g,i} \right) \\
\end{aligned}
\label{eq:dynamic_omega}
\end{equation}
\begin{equation}
\begin{aligned}
    \dot{\bm R} &=  [\bm \omega]_\times \bm R\,
\end{aligned}
\label{eq:dynamic_R}
\end{equation}
where $\ddot{\bm p} \in \mathbb{R}^{3}$ is body linear acceleration and $\bm \omega \in \mathbb{R}^{3}$ is the body angular velocity, both defined in the world frame. Both length vectors $\bm r_i \in \mathbb{R}^{3}$ and $\bm d_i \in \mathbb{R}^{3}$ are updated in real-time by using Pinocchio \cite{carpentier2019pinocchio}.

The time derivative of the Euler angles $\dot{\boldsymbol{\theta}}$ is related to the robot body angular velocity $\boldsymbol{\omega}$ through the following relationship:
\begin{equation}
\begin{aligned}
\left[\begin{array}{c}
\dot{\theta}_r \\
\dot{\theta}_p \\
\dot{\theta}_y
\end{array}\right]=\left[\begin{array}{ccc}
\text{cos}_y / \text{cos}_p & \text{sin}_y / \text{cos}_p & 0 \\
-\text{sin}_y & \text{cos}_y & 0 \\
\left(\text{cos}_y \text{sin}_p\right) / \text{cos}_p & \left(\text{sin}_y \text{sin}_p\right) / \text{cos}_p & 1
\end{array}\right] \boldsymbol{\omega}
\end{aligned}
\label{eq:thetad-omega}
\end{equation}
where under typical walking conditions, the robot never heads upward and maneuvers under only small roll and pitch angles. Therefore, Eq.~\ref{eq:thetad-omega} can be approximated as according to \cite{di_carlo_dynamic_2018}:
\begin{equation}
\begin{aligned}
\left[\begin{array}{c}
\dot{\theta}_r \\
\dot{\theta}_p \\
\dot{\theta}_y
\end{array}\right]\approx\left[\begin{array}{ccc}
\text{cos}_y  & \text{sin}_y  & 0 \\
-\text{sin}_y & \text{cos}_y  & 0 \\
 0 & 0 & 1
\end{array}\right] \boldsymbol{\omega} \approx \mathbf{R}_z^{\top}\boldsymbol{\omega}
\end{aligned}
\label{eq:thetad-omega-appox}
\end{equation}
and Eq.~\ref{eq:dynamic_omega} can be approximated as
\begin{equation}
\begin{aligned}
\frac{\mathrm{d}}{\mathrm{d} t}(\mathbf{I} \boldsymbol{\omega}) \approx \mathbf{I} \dot{\boldsymbol{\omega}} \approx \mathbf{R}_z^\mathcal{B} \mathbf{I} \mathbf{R}_z^{\top}\dot{\boldsymbol{\omega}}
\end{aligned}
\label{eq:I-omega}
\end{equation}
Let $\bm x = [\bm \theta^\top, \bm p^\top, \dot{\bm \omega}^\top, \dot{\bm p}^\top] ^\top$ be the state vector, and $\bm u = [\bm u_{g,1}^\top, \dots, \bm u_{g,4}^\top,u_{t,1}, \dots, u_{t,4}]^\top$ be the control input, where $\bm u_{g,i} \in \mathbb{R}^{3}$ is ground reaction force for leg $i$, $\bm u_{t,i} \in \mathbb{R}$ is the thrust force for $i$th thruster. Then the robot dynamic can be simplified into a linear form:
\begin{equation}
\begin{aligned}
    \bm x &= \bm A\, \bm x + \bm B \bm u + \bm h_g
\end{aligned}
\label{eq:dynamic_eom_simple}
\end{equation}
where $\bm h_g = [\mathbf{0}, \mathbf{0}, \mathbf{0}, \mathbf{g}]^\top$ and
\[
\bm{A} = \left[
\begin{array}{cccc}
\mathbf{0}_3 & \mathbf{0}_3 & \mathbf{R}_z^{\top} & \mathbf{0}_3 \\
\mathbf{0}_3 & \mathbf{0}_3 & \mathbf{0}_3 & \mathbf{1}_3 \\
\mathbf{0}_3 & \mathbf{0}_3 & \mathbf{0}_3 & \mathbf{0}_3 \\
\mathbf{0}_3 & \mathbf{0}_3 & \mathbf{0}_3 & \mathbf{0}_3
\end{array}
\right]
\]

\[\arraycolsep=1.3pt
\bm{B} = \left[
\begin{array}{cccccc}
\mathbf{0}_3 & \mycdots & \mathbf{0}_3 & \mathbf{0}_3 & \mycdots  & \mathbf{0}_3 \\
\mathbf{0}_3 & \mycdots & \mathbf{0}_3 & \mathbf{0}_3 & \mycdots  & \mathbf{0}_3\\
\mathbf{I}^{-1}\left[\mathbf{p}_1\right]_{\times} & \mycdots  & \mathbf{I}^{-1}\left[\mathbf{p}_4\right]_{\times} & \mathbf{I}^{-1}\left[\mathbf{r}_1\right]_{\times}\hat{\mathbf{e}}_1 & \mycdots &  \mathbf{I}^{-1}\left[\mathbf{r}_4\right]_{\times}\hat{\mathbf{e}}_4 \\
\nicefrac{\mathbf{1}_3}{m} & \mycdots & \nicefrac{\mathbf{1}_3}{m} &  \nicefrac{\hat{\mathbf{e}}_1}{m} & \mycdots &  \nicefrac{\hat{\mathbf{e}}_4}{m} 
\end{array}
\right]
\]
where $\left[\mathbf{\ast}\right]_{\times}$ is the skew-symmetric matrix, $\mathbf{0}_n$ and $\mathbf{1}_n$ donates $n$ by $n$ zero and identity matrix.

\subsection{Joint State Inputs}

Since the Husky $\beta$ robot is equipped with parallel-structured legs with a closed kinematic chain (as shown in Fig \ref{fig:hardware}), the link not directly connected to the knee joint was removed for kinematic modeling. Instead, a constrained ankle joint was used to mimic the knee movement, with this constraint enforced by the parallel-structured legs in the hardware. 
Pinocchio is used to calculate forward kinematics and obtain the joint Jacobian. A Jacobian-based method is then employed to compute the robot's inverse kinematics.

A hybrid joint state input is used for low-level control:
\begin{equation}
\tau_{i, \text{cmd}} = \mathbf{K}_{p, i} (q_{d, i} - q_i) + \mathbf{K}_{d, i} (\dot{q}_{d, i} - \dot{q}_i) +  \tau_{d,i}
\end{equation}
where a PD controller is used for the joint to track the desired position, and $\tau_d$ is primarily contributed during the stance phase, as calculated by
\begin{equation}
\mathbf{\tau}_j = \mathbf{J}_j^\top  \bm u_{g,j}
\end{equation}
where $\mathbf{\tau}_j \in \mathbb{R}^{4}$, including a mimic joint, and \(\mathbf{J}_j \in \mathbb{R}^{3 \times 4}\) are the joint torque vector and the foot Jacobian for $j$th leg, respectively. Additionally, \(\bm u_{g,j}\) is the vector of ground reaction forces calculated from the MPC discussed next.

\section{Model Predictive Control}
\label{sec:control}

\begin{figure*}[t]
  \centering
  \subfigure[Disturbance rejection with thruster assisted.]{\label{fig:push-w-thrust-snapshot}\includegraphics[width=\textwidth]{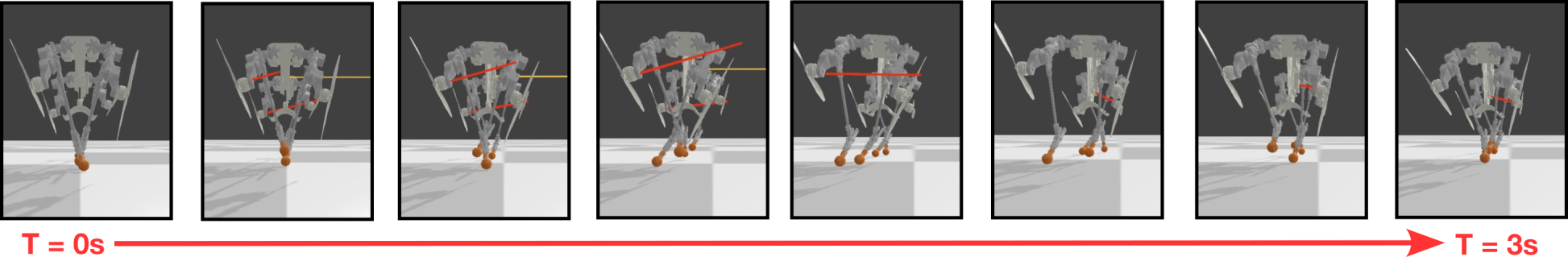}}
  \subfigure[Disturbance rejection without thruster assisted.]{\label{fig:push-wo-thrust-snapshot}\includegraphics[width=\textwidth]{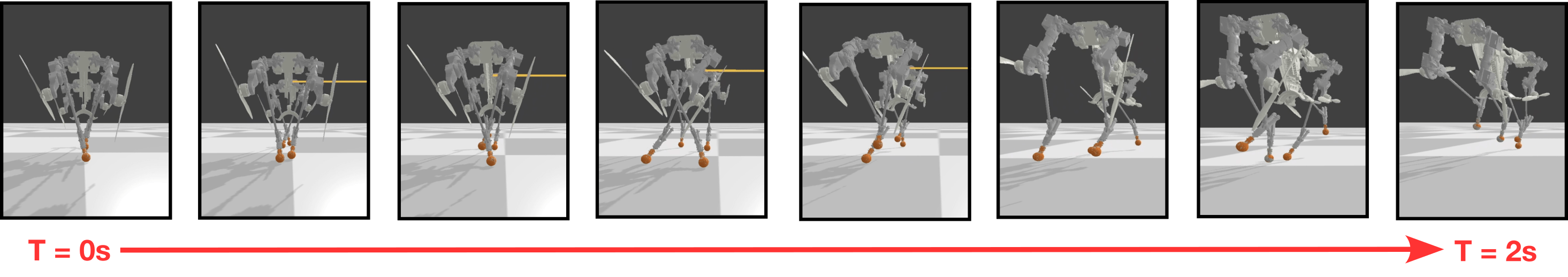}}
  \caption{Simulating a trotting gait with and without thrust force, a 40N disturbance (indicated by a yellow line) was applied from t=1s to t=1.5s. (a) With thrust forces activated to counteract the external disturbance (indicated by the red line), the robot successfully stabilized itself by t=3s. (b) Without thrust force assistance, the locomotion failed by t=2s.}
  \label{fig:push-snapshot}
\end{figure*}

\begin{figure*}[t]
    \centering
      \subfigure[Robot's CoM position comparison.]{\label{fig:push-w-thrust}\includegraphics[width=0.47\linewidth]{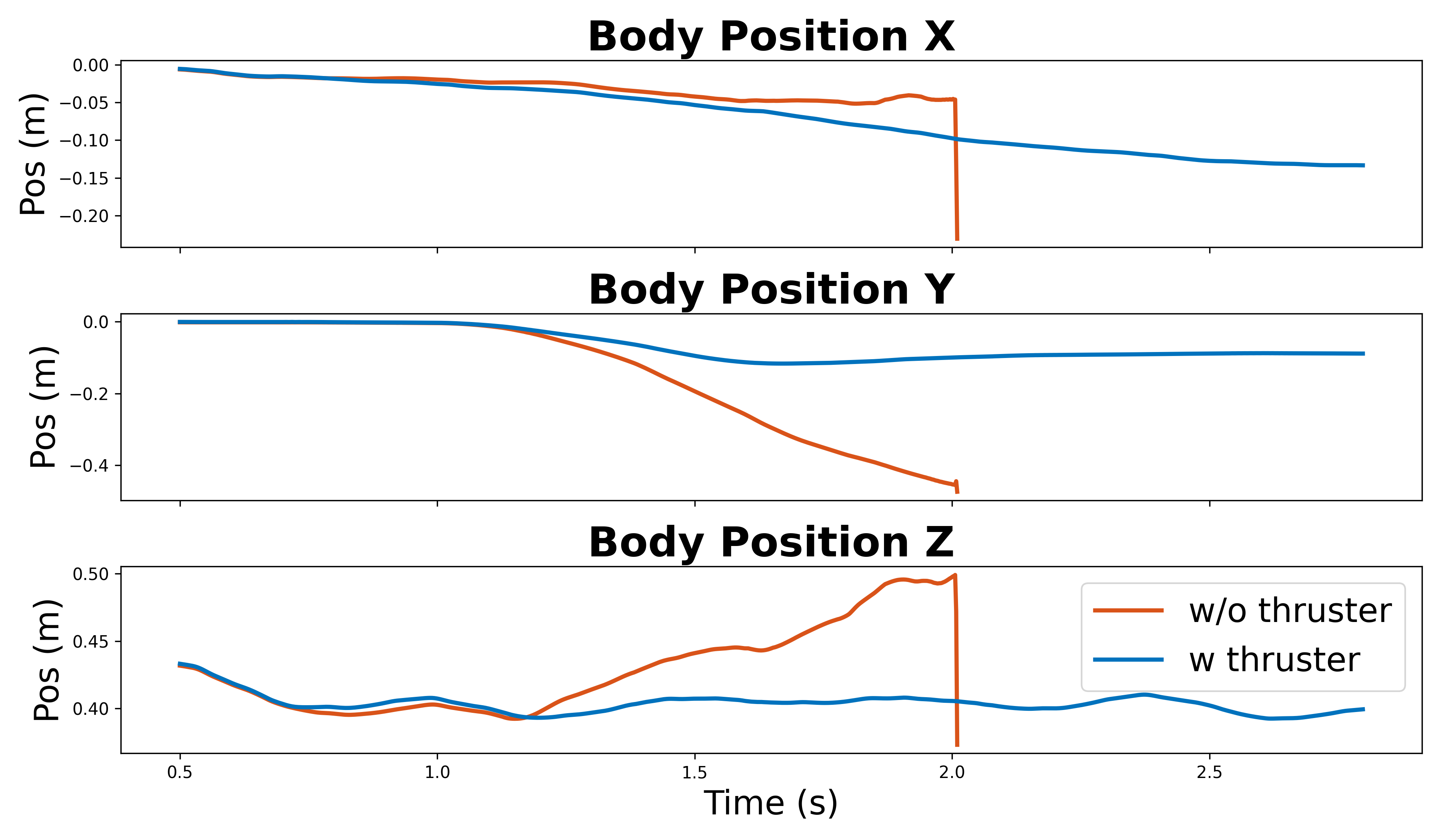}}
      \hspace{0.03\linewidth}
      \subfigure[Robot's CoM orientation comparison.]{\label{fig:push-wo-thrust}\includegraphics[width=0.47\linewidth]{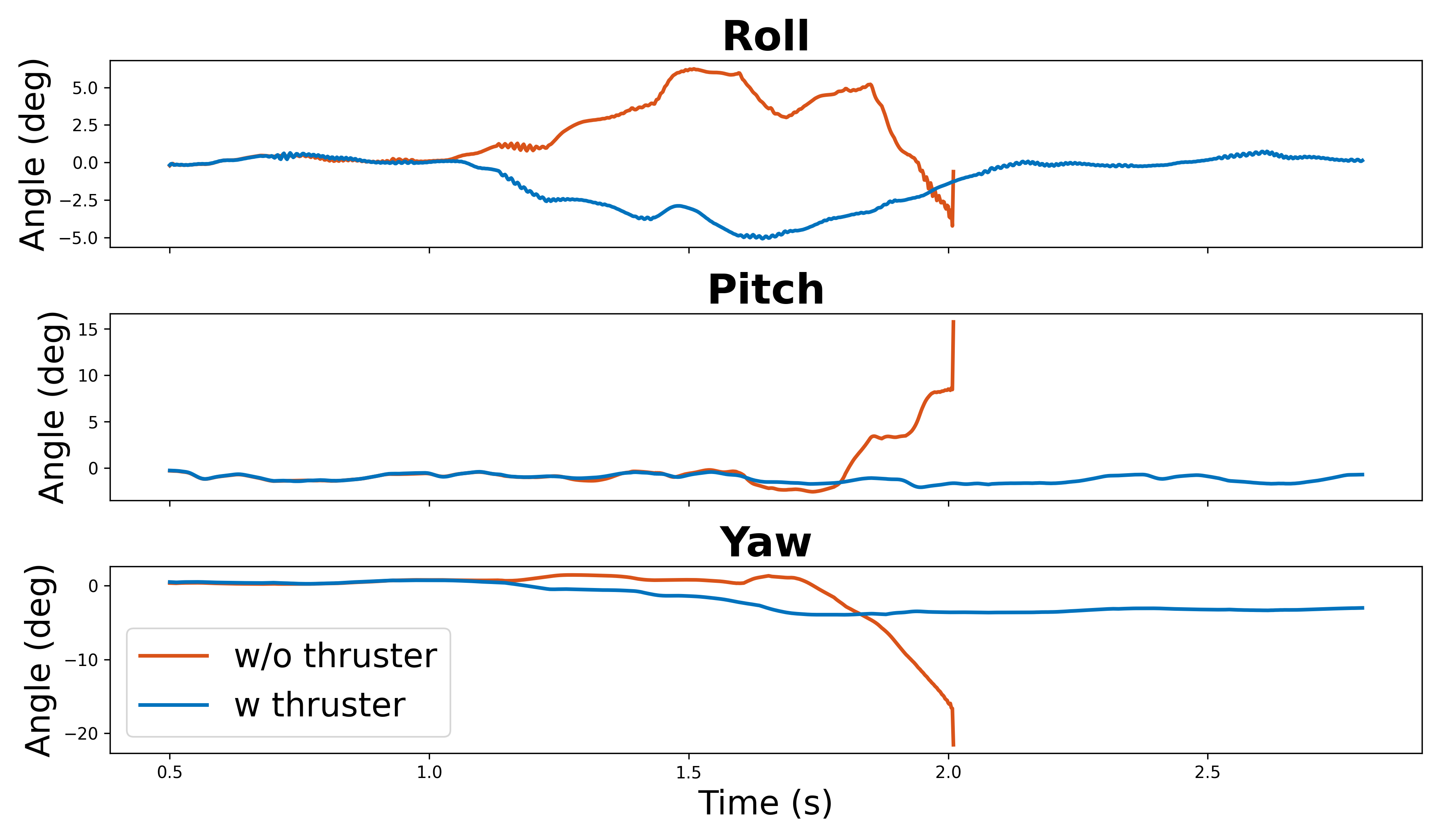}}
    \caption{Comparison of the robot's center of mass position and orientation with and without thruster activation during external disturbance.}
    \label{fig:result-push-comparsion}
\end{figure*}

In this section, we employ model predictive control (MPC) to find thruster actions. The locomotion is divided into two phases: stance and swing. The swing phase focuses on determining foot placement to move the center of mass at the desired speed. A Raibert heuristic \cite{raibert1986legged} is used in this work to find the desired foot-end location for the swing leg:
\begin{equation}
p_{d,i} = p_{ref,i} + \frac{v T_s}{2} + k (v_i - v_{d,i})
\end{equation}
where $p_{ref,i}$ is nominal foot position of foot $i$ with respect to the hip $i$ projected onto the ground plane, $T_s$ is the time duration for stance, and $v \in \mathbb{R}^{3}$ center of mass velocity in world frame. 

A fourth-order Bézier curve is used to generate the swing trajectory, which has identical first two and last two control points, ensuring zero velocity at both lift-off and touchdown.

The forces applied to the dynamical system ($\bm u$) are found at each time step using MPC and QP solver. Let $\bm x_k = [\bm \theta^\top, \bm p^\top, \dot{\bm \omega}^\top, \dot{\bm p}^\top] ^\top$ be the dynamical states at time step $k$. The dynamic models are linearized and discretized to be used in the QP solver. 

The system shown in \eqref{eq:dynamic_eom_simple} can be linearized to be used in the prediction model into the following form:
\begin{equation}
\begin{aligned}
    \bm x_{k+1} &= \bm x_{k} + \bm f_x(\bm x_k, \bm u_k)\, \Delta t \\
    &= \bm A_k\, \bm x_k + \bm B_k\, \bm u_k + \bm h_g\\
\end{aligned}
\label{eq:linearized_eom}
\end{equation}
where $\Delta t$ is the optimization prediction model time step. The optimization problem for the QP solver with $n_h$ horizon window can be defined as follows :
\begin{equation}
\begin{aligned}
    \min_{\bm u_k} \quad & {\sum_{k=1}^{n_h}} ( \bm x_k - \bm x_{r,k} )^\top \bm Q \, ( \bm x_k - \bm x_{r,k} ) +  \bm u_k^\top \bm R\, \bm u_k \\
    \textrm{s.t.} \quad 
    & \bm x_{k+1} = \bm A_k\, \bm x_k + \bm B_k\, \bm u_k + \bm h_g\\
    & u_{t,i} < u_{t}^{max} , \quad \quad  \quad \quad 
    0 < u_{t,i} \\
    & \text{for } i \in \mathcal{S}_t, \quad \quad \quad \quad \quad
    -u_{g,i,z} < 0 \\
    & u_{g,i,x} < \mu_s\, u_{g,i,z}, \quad \quad 
    -u_{g,i,x} < \mu_s\, u_{g,i,z} \\
    & u_{g,i,y} < \mu_s\, u_{g,i,z}, \quad \quad 
    -u_{g,i,y} < \mu_s\, u_{g,i,z}    
\end{aligned}
\label{eq:mpc_formulation}
\end{equation}
where $\mathcal{S}_t$ is the set of stance leg index, $\bm Q$ and $\bm R$ are cost weighting matrices, and $\bm x_{r}$ is the reference states. The friction cone constraint is linearized as shown in \eqref{eq:mpc_formulation}, where $\mu_s$ is the friction cone constraint. The thrust force range has been limited for generating single-direction solutions.

\section{Simulation Results and Discussion}
\label{sec:result}



\begin{figure*}[t]
    \centering
    \includegraphics[width = \linewidth]{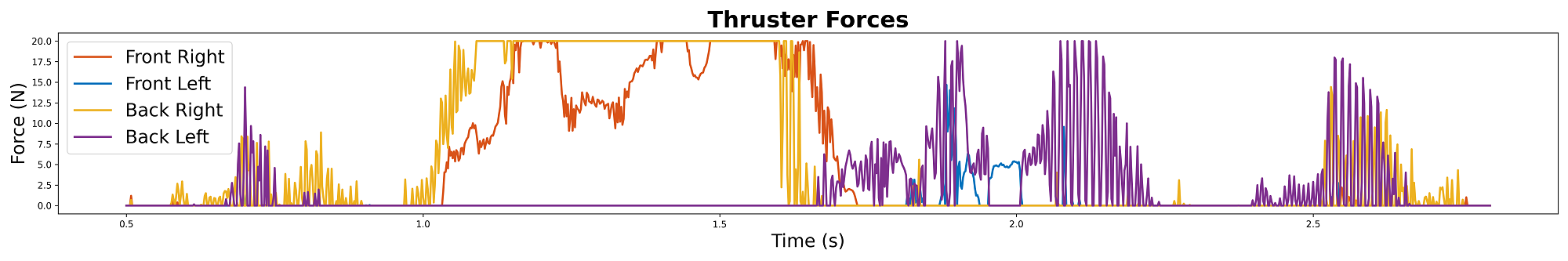}
    \caption{Thruster force during the external disturbance. From t=1.0s to t=1.7s, the right pair of thrusters activated to counterbalance the disturbance, also satisfying the 20N inequality constraint. From t=1.7s to t=2.8s, the left pair of thrusters activated to counteract the previous effects.}
    \label{fig:result-thruster-push-recovery}
\end{figure*}

\begin{figure}[t]
    \centering
    \includegraphics[width = \linewidth]{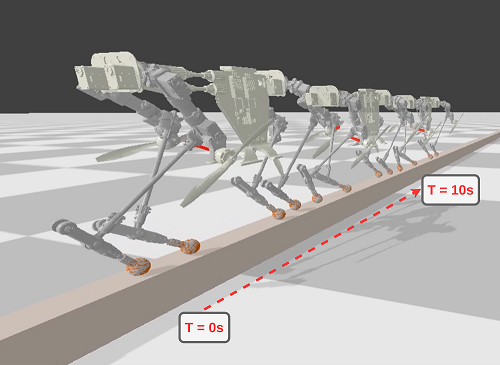}
    \caption{Snapshots of the simulation of the robot performing narrow path walking for 10 seconds.}
    \label{fig:sim-npw-snapshot}
\end{figure}

\begin{figure}[t]
    \centering
    \includegraphics[width = \linewidth]{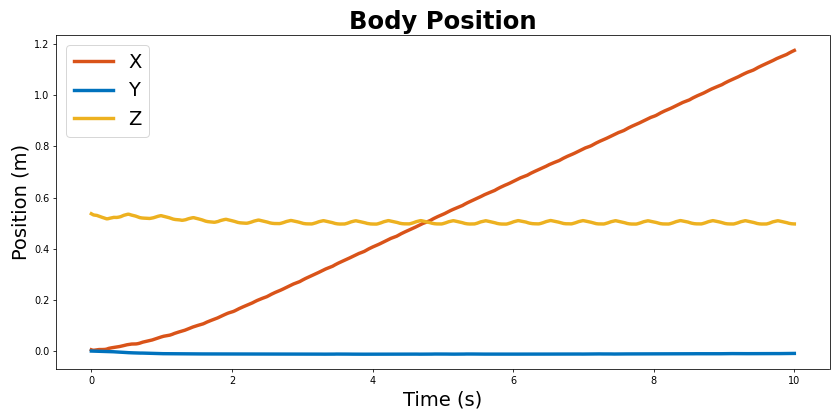}
    \caption{Plot of the robot's position state throughout the simulation. The robot has a stable lateral position and stance height while walking forward.}
    \label{fig:result-pos}
\end{figure}

\begin{figure}[t]
    \centering
    \includegraphics[width = \linewidth]{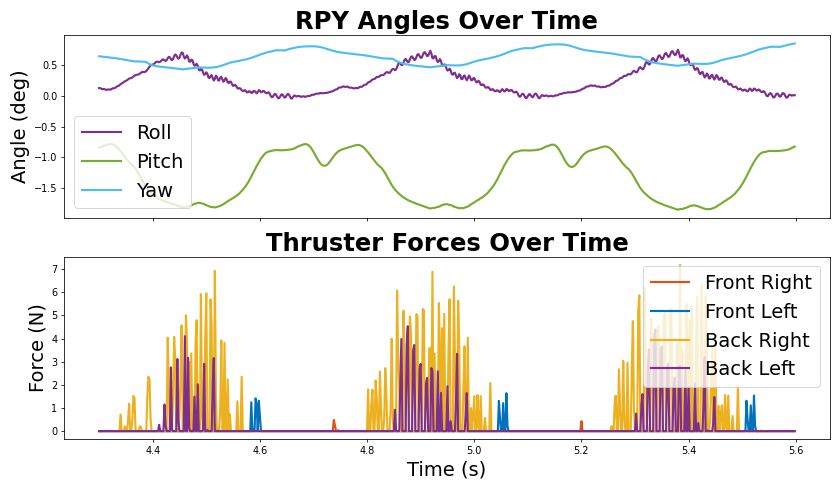}
    \caption{Plot showing the robot's attitude and thruster forces during the simulation.}
    \label{fig:result-rpyvsthrust}
\end{figure}

\begin{figure}[t]
    \centering
    \includegraphics[width = 0.95\linewidth]{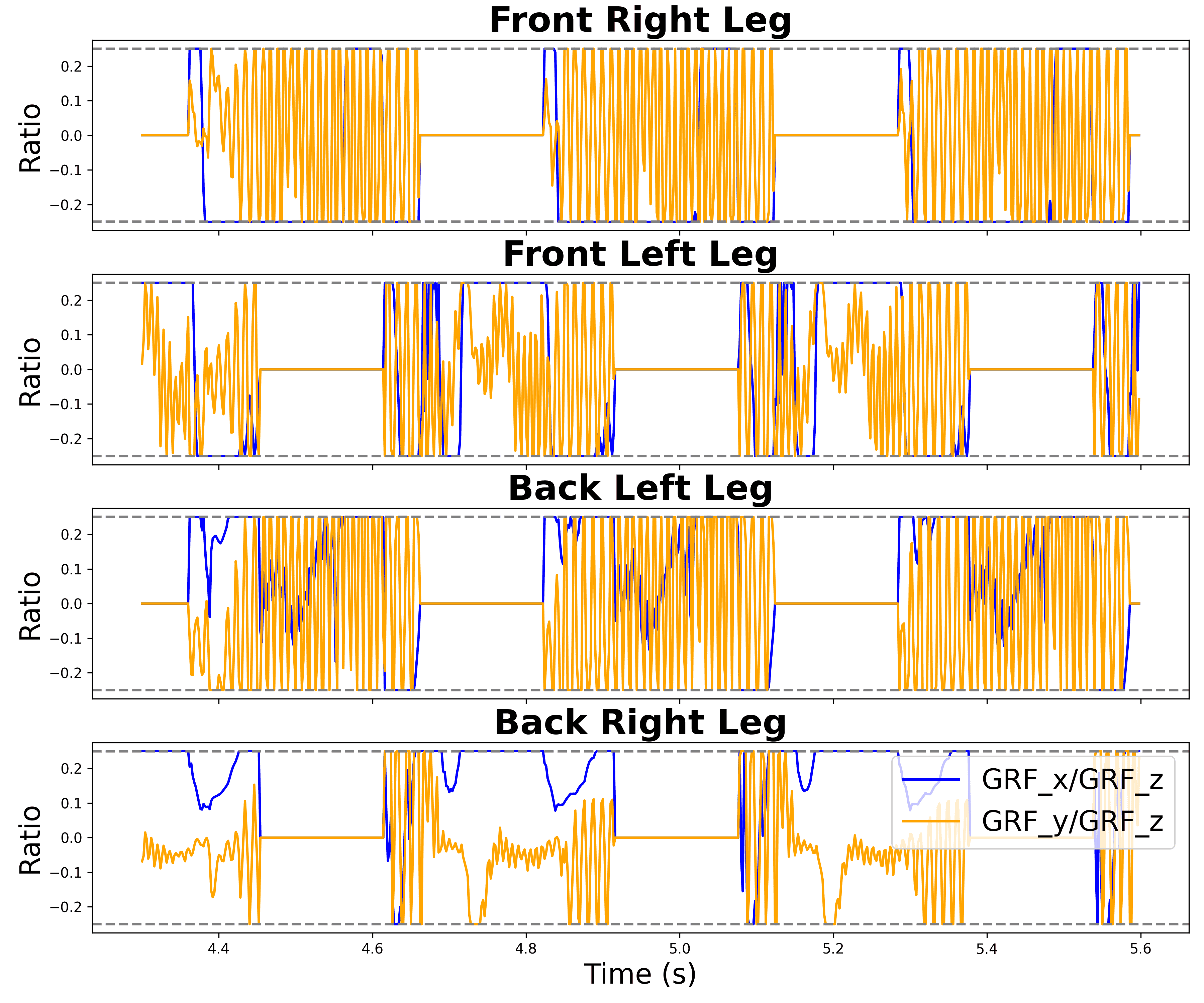}
    \caption{Plot showing the ratio of friction and normal forces on each leg. The dotted lines represent the no-slip ground force constraint limits.}
    \label{fig:result-grf-ratio}
\end{figure}

\subsection{Simulation Setup}

A PyBullet simulator \cite{coumans2016pybullet} with the robot's high-fidelity model was used for the experiments. The MPC problem state in Equation \ref{eq:mpc_formulation} in horizon of 5 was solved using the open-source QP solver quadprog \cite{quadprog2021} in 100Hz. According to \cite{qpbenchmark2024}, quadprog is one of the fastest off-the-shelf QP solvers available in Python.

\subsection{Push Recovery}

In the push recovery test, a continuous force of 40N was applied to the robot's center of mass for 0.5 seconds during a narrow path trotting gait, which is highly unstable compared to a normal wide trotting gait. The force was applied from t=1s to t=1.5s (see Figure \ref{fig:push-snapshot}). Without thruster assistance, the robot failed to recover after a counterbalance gait at 2s. However, with side propeller assistance, the MPC was able to generate a high opposite thruster force to counterbalance the external disturbance (as shown in snapshots 2-5 in Figure \ref{fig:push-w-thrust}), followed by a small same-direction force to re-stabilize the gait (as shown in snapshots 6-8 in Figure \ref{fig:push-wo-thrust}).

The comparison plots for push recovery with and without propellers can be seen in Figures~\ref{fig:result-push-comparsion}. These plots show that the robot is unstable without the use of thrusters when subjected to lateral disturbance during narrow-path walking. The thruster forces used to counteract this disturbance can be seen in Figure~\ref{fig:result-thruster-push-recovery}, showing a large initial force contribution from the robot's right propellers to recover from the disturbance.

\subsection{Beam walking}

The previous push recovery test demonstrated the roll dynamic stability of the proposed controller. To further showcase the robot's capabilities utilizing such roll stability, a narrow beam walking test was conducted. In this test,  a narrow beam with dimensions of 0.1m in width and 0.1m in height was set up for the robot to perform the maneuver using both ground reaction force and thrust force (see Figure \ref{fig:sim-npw-snapshot}). The simulation result plots can be seen in Figures~\ref{fig:result-pos} to \ref{fig:result-grf-ratio}, showing the robot's simulated position and attitude in addition to the thruster forces, and the ratio of ground friction force and normal force on each leg.

Figures \ref{fig:result-pos} and \ref{fig:result-rpyvsthrust} show that the robot's lateral position, height, and orientation are stable as the robot walks forward. The thruster forces are shown in Figure~\ref{fig:result-rpyvsthrust} where the forces do not go over 7 N, which is relatively small compared to the thruster's maximum thrust of approximately 100 N. This shows that the thrusters are used lightly to stabilize the roll dynamics during the narrow path walking. The sagittal propeller played a significant role in this maneuver, ensuring the ground friction cone constraint was satisfied throughout the entire process (see Figure \ref{fig:result-rpyvsthrust} and Figure \ref{fig:result-grf-ratio}).

\section{Conclusions and Future Work}

We have developed a simulation model in PyBullet using the robot's actual physical model and implemented an MPC-based thruster and ground force regulation controller to perform narrow-path walking and lateral disturbance rejection. We have successfully utilized the Husky $\beta$ multi-modal capability to stabilize its frontal dynamics in these simulation scenarios. These thrusters can be used to help regulate ground reaction forces to enforce the no-slip conditions and also be used to transform the robot into UAV mode and fly over obstacles or rough terrains. This shows that our design can expand the robot's stability and locomotion plasticity. 

For future work, we can attempt to implement this controller and perform the same experiments on the actual robot. This is significantly more challenging as the actual robot can have limitations due to computing power, noisy measurements, communication delays, etc. Overcoming these limitations might require iterative work or the need to develop a robust controller and estimation algorithm.

\section{ACKNOWLEDGMENT}
This work is supported by the Technology Innovation Institute (TII), UAE, and the U.S. National Science Foundation (NSF) CAREER Award (Award No. 2340278).










\printbibliography

\end{document}